%% file: PRICAI.tex
\documentclass[runningheads]{llncs}
\usepackage{graphicx}
\usepackage{float}
\usepackage[english,vietnamese]{babel}
\usepackage{multirow}
\usepackage{booktabs}
\usepackage{amsmath}
\usepackage{arydshln}
\usepackage{amssymb}
\usepackage{xspace}
\usepackage[hidelinks]{hyperref}
\usepackage{xcolor}
\usepackage[misc]{ifsym}
\hypersetup{
    colorlinks,
    linkcolor={blue!80!black},
    citecolor={blue!80!black},
    urlcolor={blue!80!black}
}

\begin{document}

\title{VSEC: Transformer-based Model for Vietnamese Spelling Correction}
\input{authors}
\maketitle              % typeset the header of the contribution

\selectlanguage{english}
\input{abstract}

\input{Sections/intro}
\input{Sections/related_work}
\input{Sections/our_approach}

\input{Sections/experiment}
\input{Sections/conclusions}

\subsubsection*{Acknowledgement.}
This work has been supported by Vietnam National University, Hanoi (VNU), under Project No. QG.18.61.

\bibliographystyle{splncs04}
\bibliography{references}

\end{document}

%% file: authors.tex
\author{Dinh-Truong Do \inst{1}\textsuperscript{(\Letter)} \and Ha Thanh Nguyen\inst{2}\and
Thang Ngoc Bui\inst{1}, 
\\and Hieu Dinh Vo\inst{1}}
%\orcidID{0000-1111-2222-3333}
\authorrunning{T. Do et al.}
\institute{VNU University of Engineering and Technology, Hanoi, Vietnam \and
Japan Advanced Institute of Science and Technology, Ishikawa, Japan \\
\email{17021090@vnu.edu.vn}, \email{nguyenhathanh@jaist.ac.jp}, \\\email{thangbn@vnu.edu.vn},  \email{hieuvd@vnu.edu.vn}}

% \author{}
% \institute{}
\def\instnum{}

%% file: abstract.tex
\begin{abstract}

Spelling error correction is one of topics which have a long history in natural language processing. Although previous studies have achieved remarkable results, challenges still exist. In the Vietnamese language, a state-of-the-art method for the task infers a syllable's context from its adjacent syllables. The method's accuracy can be unsatisfactory, however, because the model may lose the context if two (or more) spelling mistakes stand near each other. In this paper, we propose a novel method to correct Vietnamese spelling errors. We tackle the problems of mistyped errors and misspelled errors by using a deep learning model. The embedding layer, in particular, is powered by the byte pair encoding technique. The sequence to sequence model based on the Transformer architecture makes our approach different from the previous works on the same problem. In the experiment, we train the model with a large synthetic dataset, which is randomly introduced spelling errors. We test the performance of the proposed method using a realistic dataset. This dataset contains 11,202 human-made misspellings in 9,341 different Vietnamese sentences. The experimental results show that our method achieves encouraging performance with 86.8\% errors detected and 81.5\% errors corrected, which improves the state-of-the-art approach 5.6\% and 2.2\%, respectively.

\keywords{Vietnamese spell correction  \and deep learning \and subword level \and Vietnamese realistic dataset.}
\end{abstract}

%% file: Sections/intro.tex
\section{Introduction}
Spelling error correction~\cite{church1991probability} is an important task, which aims to detect and correct spelling errors in a document. It is used for a variety of natural language applications, including search queries~\cite{li2006exploring,ahmad2005learning,gao2010large}, message filtering systems~\cite{gong2019context,wint2017spell,wint2018non}, and optical character recognition (OCR)~\cite{takahashi1990spelling,taghva2001ocrspell,reynaert2011character}. In this paper, we consider Vietnamese spelling correction in general.

% The traditional methods of using statistical models for spelling error correction have had certain limitations in terms of grasping the context of sentences. Therefore, models using machine learning techniques are an appropriate approach to research and apply. 

In most cases, there are two kinds of errors in the Vietnamese language: mistyped errors and misspelled errors~\cite{nguyen2019deep}. Mistyped errors are errors that occur during the typing process. The majority of these mistakes are caused by the typist's unintentional actions, such as pressing the wrong key between two adjacent characters on the keyboard. Furthermore, these errors typically stop at the syllable level, and they can be detected if the typist carefully reviews the text. Mistyped errors can be classified into two smaller categories: non-word errors and real-word errors. Non-word errors mean that the words completely do not exist in the dictionary. Real-word errors, on the other hand, are errors that the words that are still in the dictionary but used in the wrong contexts.

A misspelled error is one that the typists did not realize was incorrect. This type of error is caused by regional pronunciation mistakes or the difficulty of some Vietnamese words. Compared to mistyped errors, misspelled errors are harder to detect since we not only need to rely on the context but also have knowledge of the standard dialect to detect these errors. Table~\ref{tab:tab1} shows some examples of spelling errors in Vietnamese.

\begin{table}
\centering
\caption{Examples of Vietnamese spelling errors.}
\label{tab:tab1}
  \selectlanguage{vietnamese}
\setlength{\tabcolsep}{0.5em} % for the horizontal padding
{\renewcommand{\arraystretch}{1.3}% for the vertical padding
\begin{tabular}{|c|@{}c@{}|}\hline
Non-word mistyped
&
\begin{tabular}{c}
\hspace{0.7cm} Original: \textbf{Trới} hôm nay đẹp quá.\hspace{0.7cm}  \\
Today's\textbf{ ??? }is so beautiful.  \\\hdashline
Correct: \textbf{Trời} hôm nay đẹp quá  \\
Today's \textbf{weather} is so beautiful.  \\
\end{tabular}
\tabularnewline\hline
Real-word mistyped
&
\begin{tabular}{c}
\hspace{0.6cm} Original: \textbf{Sướng} còn đọng trên lá.\hspace{0.6cm}  \\
\textbf{Pleasure} remains on the leaves.  \\\hdashline
Correct: \textbf{Sương} còn đọng trên lá. \\
\textbf{Dew lingers} remains on the leaves  \\
\end{tabular}
\tabularnewline\hline
Misspelled
&
\begin{tabular}{c}
Original: Thuyền của tôi đang \textbf{leo} trong bến.  \\
My boat \textbf{is climbing} in the dock  \\\hdashline
Correct: Thuyền của tôi đang \textbf{neo} trong bến.  \\
My boat \textbf{is anchored} in the dock.  \\
\end{tabular}
\tabularnewline\hline
\end{tabular}

}
\end{table}

\selectlanguage{english}

Spelling error correction is a problem that has received a lot of attention from the natural language processing community. In the Vietnamese language, there were a large number of studies approaching this problem by adopting statistical language models~\cite{nguyen2008vietnamese,huong2015using,nguyen2015normalization,nguyen2019ocr}, such as N-gram. These traditional models learn the context by training on a large dataset. This method, however, has its limitation: the context of a syllable can only be grasped by the adjacent syllables. For sentences that have two or more spelling mistakes next to each other, it is harder for the model to identify errors. In recent years, the application of deep learning models to the Vietnamese spelling check is a new trend that interests researchers~\cite{nguyen2019deep}. The advantage of this approach is that the context of a syllable is not constrained by surrounding syllables, allowing the model to detect spelling errors more accurately. Although some positive results have been obtained, almost all studies on this method primarily focus on correcting certain types of spelling errors, making it difficult to apply in the real world.

% due to the limitations of the deep learning model used, these approaches still have limitations in terms of speed and the amount of data used for training, making them difficult to apply in the real world.

% For the Vietnamese language, there is outstanding research \cite{huong2015using}, when training the model with N-gram method on a large dataset. Up to now, this is also the model with the best results for Vietnamese spelling correction.  

In this paper, we propose a subword-level Transformer based model for Vietnamese spelling correction and evaluate it with a realistic dataset. The contributions of the paper include:

\begin{itemize}
 \item A deep learning method for Vietnamese spelling correction, where both mistyped errors and misspelled errors are considered;
 
 \item A process of generating Vietnamese spelling errors, which artificially add errors to a non-error sentence; this process is used to produce a large number of artificial mistakes for deep learning models to learn from;

 \item A public dataset of human-made spelling errors, which includes 9,341 sentences in 4,582 different types of errors; this dataset is a benchmark for evaluating various approaches.

\end{itemize}

The rest of the paper is organized as follows. The next section briefly introduces some related works. Section~\ref{methodology} details each step of the proposed method. Section~\ref{experimental_results} presents the experimental results of the models, and we draw some conclusions in Sect.~\ref{conclusions}.

%% file: Sections/related_work.tex
\section{Related Work}

\label{related_work}

Spelling error correction is an essential part of natural language processing (NLP). In the Vietnamese language, many methods have been proposed for this problem.  Previous approaches can be primarily divided into two categories. One employs traditional statistical language models and the other uses machine learning.

In 2008, Phuong H. Nguyen et al.~\cite{nguyen2008vietnamese} proposed a statistical method that used POS Bigram (Part Of Speech Bigram) to detect suspected syllables. Minimum Edit Distance and SoundEx algorithms have been applied to generate suggestion candidates in the correcting phase. To rank these candidates, some heuristics in relevant criteria are also used.

Nguyen Thi Xuan Huong et al.~\cite{huong2015using} developed an N-gram language model for Vietnamese spell correction. A large unlabeled dataset is used to learn the context of syllables. Specifically, the N-gram score for each syllable in the candidate set is calculated based on the frequency of occurrence in unigram, bigram, and trigram. The model creates the candidate set based on changing characters in syllables corresponding to typing errors, consonant errors, etc. The current syllable is considered an error if a syllable in the candidate set has a higher N-gram score than the current one.
 This approach is currently state-of-the-art with approximately 94\% F1 score on their experimental data.
 
 By detecting and correcting spelling errors, Nguyen Hong Vu et al.~\cite{nguyen2015normalization} proposed a method to normalize Vietnamese tweets. The words with spelling errors were detected based on a dictionary. The model corrected the errors by combining the Vietnamese vocabulary structure with a language model based on improved Dice and SRILM (A language model).

Spelling error correction is an important step in improving the accuracy of OCR-generated text. For Vietnamese OCR errors, Quoc-Dung Nguyen et al.~\cite{nguyen2019ocr} developed an approach for generating and scoring correction candidates based on linguistic features. The spelling errors will be detected based on the unigram, bigram, and trigram dictionaries. After the detecting phase, a candidate set for each syllable error will be generated by applying insertion, deletion, and substitution operators. The candidates with high score which is calculated based on linguistic features such as Syllable Similarity, Bigram frequency, Trigram frequency, and Edit Probability will be included in the suggestion list.

In 2018, Nguyen Ha Thanh et al.~\cite{nguyen2019deep} proposed a deep learning method to solve Vietnamese consonant misspell errors. To identify and correct error positions, the model employs misspell direction encoding and bidirectional stacked LSTM architecture.

Spelling error correction can be formulated as a problem of translating a misspelling sequence to a corrected one. This type of problem can be solved with typical methods used for machine translation. Some researchers have applied Neural Machine Translation (NMT) models~\cite{zhou2017spelling,buyuk2020context,gu2017chinese} to correct spelling errors in popular languages such as English and Chinese. Their positive results demonstrate that this is a viable solution to the problem of spelling error correction.

% https://www.mdpi.com/2079-9292/9/10/1670

One of challenges with NMT is the out-of-vocabulary problem. Increasing the model's vocabulary size is a simple way to solve this problem. However, if the vocabulary is too large, the dimension of the vector embedding will be too high. It increases the computation time and adds complexity to the model's training. To address this problem, some studies~\cite{tacorda2017controlling,choudhary2018neural} applied Byte Pair Encoding (BPE), which tokenizes sentences at subword level~\cite{sennrich2015neural}. This technique keeps input length to a reasonable level while handling unseen and rare words.

% % https://sci-hub.se/10.1109/ialp.2017.8300571

%% file: Sections/our_approach.tex
\section{Methodology}

\label{methodology}

\subsection{Problem Statement}

Vietnamese spelling correction can be formulated as follows. Given a set of syllable sequences \textit{X} = $\{x^i=(x^i_1, x^i_2 ..., x^i_n) \}$ with some errors in $x^i$, and a set of syllable sequences \textit{Y} = $\{y^i=(y^i_1, y^i_2 ..., y^i_m) \}$, where $y^i$ is error-free. The goal is to transform each sequence $x^i$ into corresponding sequence $y^i$. Following that, the task can be considered as a problem of  learning a
function $\textit{f} : \textit{X} \rightarrow \textit{Y}$ that satisfies $\textit{f}(x^i) = y^i$.

\subsection{Model Overview}

The state-of-the-art method for Vietnamese spelling correction is to use a statistical language model. Although the method is trainable on a large dataset, it uses a limited context. This motivates us to create a new model  which exploits broad context.

\begin{figure}[H]
  \begin{center}
  \includegraphics[width=1\textwidth]{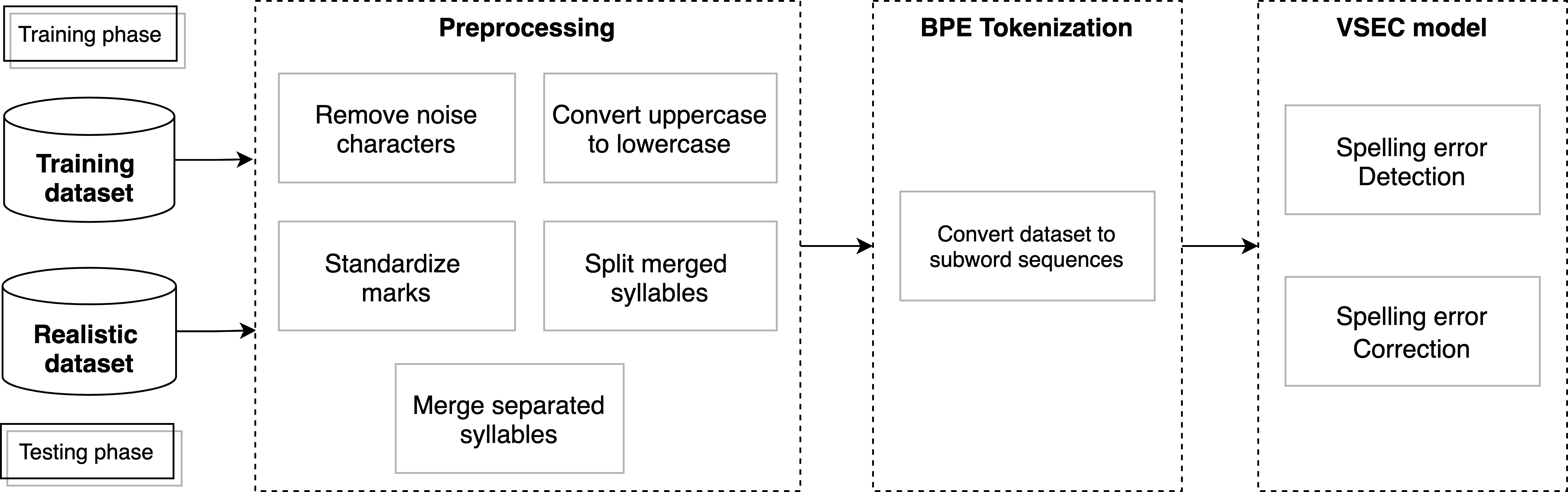}
  \end{center}
  \caption{Model pipeline}
  \label{fig:f.1}
\end{figure}

Figure~\ref{fig:f.1} illustrates an overview of our proposed model, called VSEC, in training and testing processes. The pipeline of VSEC is composed of three components: a preprocessing module, a tokenization module, and a Transformer-based model. The original data is first preprocessed to remove any noise that might appear in the sentence. The BPE tokenizer then converts each sentence into a sequence of tokens. Finally, the Data Loader feeds the sequence into the Transformer-based model for training.

\subsection{Preprocessing}

To ensure that the results are reliable, data quality assurance is critical. Each input sequence needs to be removed noise before proceeding to the next phase. Our preprocessing module, in particular, consists of five steps:

\begin {itemize}
 \item \textbf{Step 1: Remove noise characters} - In this step, we remove characters that are not  useful for learning the context of a sentence such as emojis, line break characters.

 \item \textbf {Step 2: Convert uppercase to lowercase} - Using both upper and lower case can affect the model's data density. Therefore, all uppercase characters are converted to lowercase.

 \item \textbf {Step 3: Standardize marks} -  In this step, each syllable is converted to the syllable in the telex typing form. Following that, the data is standardized using a mapping set between the telex syllable and the correct mark syllable.
 
  \selectlanguage{vietnamese}
 \textit{For example: Syllable ``cuả'' => Telex syllable  ``cuar''  => Standardized syllable ``của''}.

  \selectlanguage{english}
 \item \textbf {Step 4: Split merged syllables} - The appearance of merged syllables in the dataset may have a detrimental effect on the model's learning.  We use the Peter Norvig word segmentation algorithm to solve this problem~\cite{norvig2009natural}. When syllables are merged in the Vietnamese language, they transform into the syllables in the telex typing form. This occurs as a result of the Vietnamese typing tools' mechanism. Therefore, while calculating probability using the Peter Norvig algorithm, it is critical to convert the telex syllable to the standard Vietnamese syllable.
 
  \item \textbf {Step 5: Merge separated syllables} - A syllable that has spaces between its characters is known as a separated syllable. The model is also more difficult to converge due to the appearance of these strange syllables. To solve this problem, we employ the Trie structure~\cite{thue1912uber}, which has demonstrated its ability to browse prefixes.
  
\end {itemize}

\subsection{Tokenization}

In Vietnamese, each space-separated token is in monosyllabic form. Therefore, we call the word level in English as the syllable level in Vietnamese from now on. At first glance, using syllable level as the input seems like a good idea. However, this level is not well suited for spelling error correction, as we can have difficulties with misspelling syllables or rare syllables (out of vocabulary). It makes the model harder to learn the sentence's context. One of the solutions to this problem is to use the character level. Nevertheless, breaking syllables into characters will increase the sequence length. As a result, the model is large and slow to converge.

Subword level is between syllable level and character level. It keeps the input length at a reasonable level while addressing the out-of-vocabulary problem. For example, we can split a Vietnamese misspelling syllable “nghành” into two tokens: “ngh” and “ành”, and present “nghành” by vectors of these tokens. The BPE algorithm is used to construct a subword dictionary~\cite{sennrich2015neural}. Given a large corpus, this tokenization technique groups characters into frequent sequences. It is totally unsupervised and requires no information about the context of the sentence.  An example of how BPE obtains vocabulary from raw text is shown in Table~\ref{tab:tab2}.
  
\setlength{\tabcolsep}{0.5em}
\renewcommand{\arraystretch}{1.3}

\begin{table}[H]
    \centering{%
\caption{An example of how BPE obtains vocabulary given a raw sequence}
    \label{tab:tab2}
    \begin{tabular}{ccc}
    
\hline

\textbf{Iteration} & \textbf{Sequence} & \textbf{Vocabulary}\\

\hline
  0& a {} {}  t {} {}  e {} {}  /w {} a {} {} t {} {} /w& \{a, t, e, /w\}\\
  1& at {} {} e {} {}/w  {} at {} {} /w & \{a, t, e, /w, at\}\\
  2& at {} {} e {} {}/w  {} at/w & \{a, t, e, /w, at, at/w\}\\
  3& at {} {} e/w  {} at/w & \{a, t, e, /w, at, at/w, e/w\}\\
  4& ate/w  {} at/w & \{a, t, e, /w, at, at/w, e/w, ate/w\}\\
\hline
\end{tabular}}
\end{table}

\begin{figure}[H]
  \begin{center}
  \includegraphics[width=238pt]{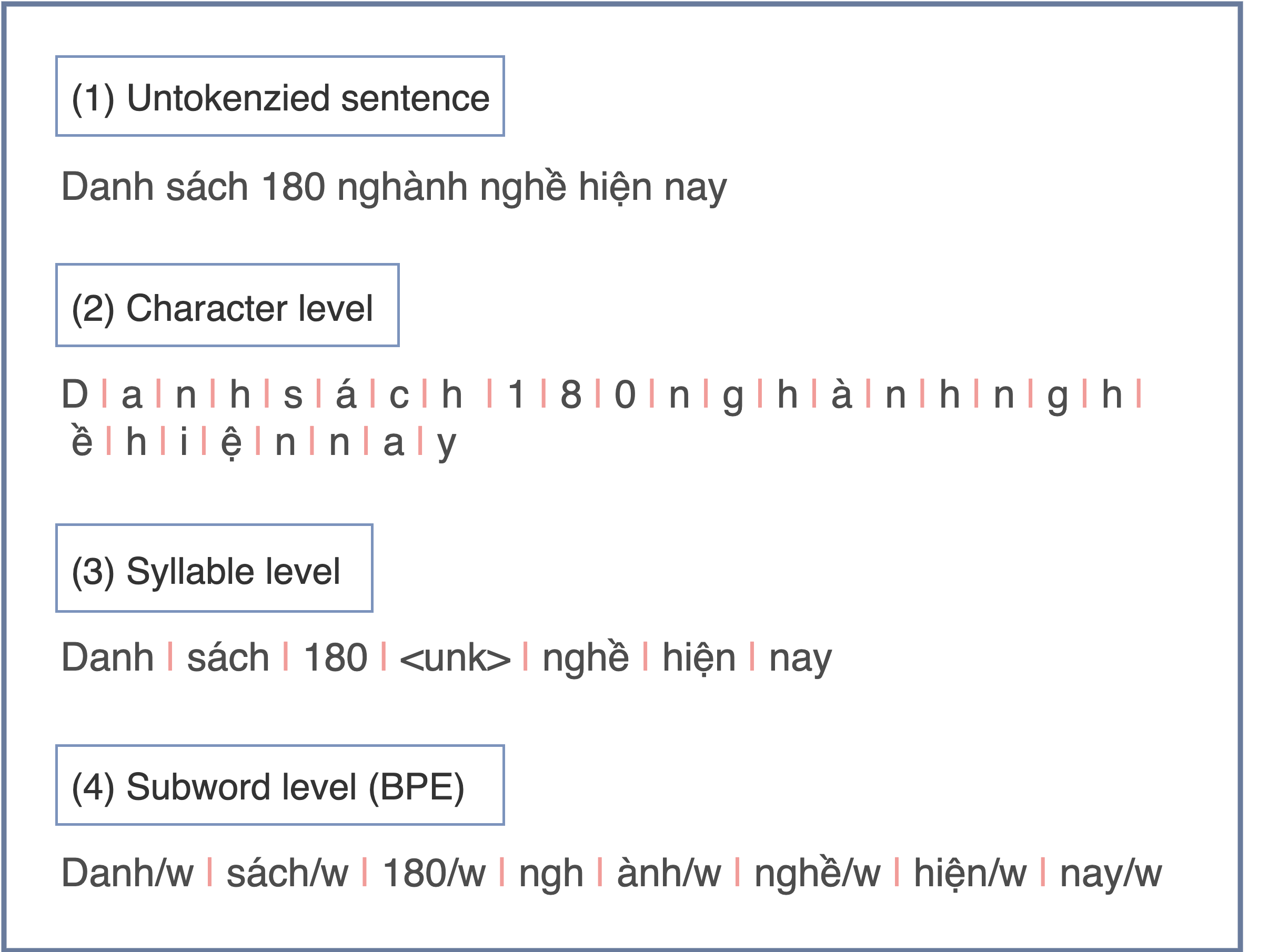}
  \end{center}
  \caption{Comparison of different tokenization levels. Tokens are separated by ``$|$''. ``nghành'' is not a Vietnamese syllable. }
  \label{fig:f2}
\end{figure}

The algorithm of BPE is as follows. Firstly, a special token $/w$ is appended to each syllable to indicate the end position of a syllable. Then, we split all sentences in the corpus into characters. At this point, the vocabulary only contains single characters. After that, we iteratively count all token pairs and merge each occurrence of the most frequent pair (Y, Z) into a new token YZ and add it to
the vocabulary. The size of the final vocabulary is equal to the total number of merge operations and initial characters. The number of merge operations is the only parameter of the BPE algorithm. We will have a large vocabulary if this number is large. An example of the BPE tokenization result is in Fig. \ref{fig:f2}.

\subsection{Transformer Model}
Based on the idea of treating Vietnamese spelling error correction as a machine translation problem, the proposed model learns to translate the sentence having spelling errors to the corrected one. Specifically, we use the Seq2seq architecture based on Transformer~\cite{vaswani2017attention} as our baseline. The Transformer encodes a misspelling sentence to a context hidden state using a stack of L encoder blocks, each of which employs a multi-head self-attention layer and a feed-forward network. The decoder uses the encoder's hidden states and the sequence of previous target tokens to generate the target hidden state by applying a stack of L decoder blocks. The decoder block has the same architecture as the encoder one, except it has an extra attention layer over the encoder’s hidden states.

    The goal of Transformer is to predict the next token \textbf{$y_t$}, given the source tokens \textbf{$(x_1, x_2, ...,x_n)$}. The formulas of this process are: 

 \begin{equation}
    \label{eqn:eq1}
    \textbf{h}^{src}_{1..n} = encoder(\textbf{E}^{src}_{x_{1..n}})
\end{equation}

 \begin{equation}
 \label{eqn:eq2}
    \textbf{h}_{t} = decoder(\textbf{E}^{trg}_{y_{1..t-1}}, \textbf{h}^{src}_{1..n})
\end{equation}

 \begin{equation}
 \label{eqn:eq3}
    \textbf{P}_{t}(w) = softmax(\textbf{W}^{T}\textbf{h}_{t})
\end{equation}

The embedding matrix is represented by $\textbf{E} \in \textbf{\text{R}}^{d \times |V|}$, where \textit{d} is the embedding dimension, and $|V|$ is the size of the vocabulary. The value of $x_i$ represents the position of the \textit{i}-th token in the vocabulary. The encoder's hidden states are denoted by $\textbf{h}^{src}_{1..n}$ and the target hidden state of the next token is denoted by $\textbf{h}_t$. After obtaining the target hidden state, the model determines the next token to be generated by feeding $\textbf{h}_t$ into the fully connected (dense) layer behind. Particularly, the fully connected layer has $|V|$ hidden units activated by the Softmax function, which produces scores whose total is 1.0. These values correspond to the generation probability distribution of the next token.

% We get the generation probability distribution of the next token by applying linear layer and softmax operation.

%% file: Sections/experiment.tex
\section{Experimental Results}

\label{experimental_results}
\subsection{Dataset}

  \selectlanguage{vietnamese}
  
To generate a dataset for training the proposed model, we created a process for artificially adding errors to non-error sentences in Vietnamese. This process is referred to as Error Generator. We began by extracting 5 million sentences from a Vietnamese news corpus\protect\footnotemark\hspace{0em}\footnotetext{\url{https://github.com/binhvq/news-corpus}}, which was crawled from several prominent Vietnamese websites. A fusion table was also constructed, in which each syllable is linked to a group of other candidates, to present common types of Vietnamese spelling errors such as mistyped errors, consonant errors. Then, at random, we selected 8\% of the syllables in the sentences to artificially generate errors, with 90\% of them being replaced with other syllables, 5\% being removed, and 5\% being duplicated. The difference between VSEC Error Generator and others is the use of add and delete operators, which represents errors when the typists often use copy and paste.

  \selectlanguage{english}

In addition, we developed a realistic dataset for testing. We sampled the contents of 618 documents at Tailieu\protect\footnotemark\hspace{0em}\footnotetext{\url{https://tailieu.vn}}, an educational material website. To ensure that the dataset includes a significant amount of incorrect sentences, we sampled documents from lower quality texts, and thus the error rate of the dataset higher. Three people handled three phases of labeling to carefully correct spelling errors in the texts. The dataset includes 9,341 sentences, which contain 11,202 spelling errors in 4,582 different types\protect\footnotemark\hspace{0em}\footnotetext{\url{https://github.com/VSEC2021/VSEC}}.

\subsection{Evaluation Metric}

We utilized syllable-level precision, recall, and F1 score which are common in the community~\cite{nguyen2008vietnamese,huong2015using,nguyen2015normalization}. In addition, we evaluated the accuracy of both detection and correction tasks. Specifically, we used six metrics:
\\
\\
\begin{equation}
    \textit{Detection Precision} = \frac{\text{\textit{\# of true detections}}}{\text{\textit{\# of error detected}}}     \ \ \ \ \ \ \ (DP)
\end{equation}

\begin{equation}
    \textit{Detection Recall} = \frac{\text{\textit{\# of true detections}}}{\text{\textit{\# of actual errors}}}      \ \ \ \ \ \ \ \ \ \ \ (DR)
\end{equation}

\begin{equation}
    \textit{Detection F1-score} = \frac{\text{\textit{2 * DR * DP}}}{\text{\textit{DR + DP}}}     \ \ \ \ \ \ \ \ \ \ \ \ \ \ \ \ (DF)
\end{equation}

\begin{equation}
    \textit{Correction Precision} = \frac{\text{\textit{\# of true corrections}}}{\text{\textit{\# of error detected}}}     \ \ \ \ (CP)
\end{equation}

\begin{equation}
    \textit{Correction Recall} = \frac{\text{\textit{\# of true corrections}}}{\text{\textit{\# of actual errors}}}     \ \ \ \ \ \ \ \ (CR)
\end{equation}

\begin{equation}
    \textit{Correction F1-score} = \frac{2 * CR * CP}{CR + CP}     \ \ \ \ \ \ \ \ \ \ \ \ \ \ \ \ (CF)
\end{equation}

\subsection{Experimental Setting}

The BPE Tokenizer is used in the experiments based on HuggingFace's library.\protect\footnotemark\hspace{0em}\footnotetext{\url{https://github.com/huggingface/tokenizers}} Specifically, the BPE Tokenizer was trained on news corpus to build a subword-level vocabulary. We set the vocab size to 30,000 and kept other default hyperparameters. 

In the training phase, we use Adam optimizer with Cross-Entropy Loss to train the neural network model with Transformer architecture. Through the experiment, the model achieved the best results with hyperparameters are shown in Table \ref{tb.t2}.

\begin{table} 
  \centering
  \caption{Parameters}
  \label{tb.t2}
  \begin{tabular}{lc}
    \hline
    \textbf{Parameters}& \textbf{Value}\\
    \hline
    Embedding dimension&512\\
    Sequence length&200\\
    Number of head in multi-head attention&8\\
    Number of encoder/decoder layers&3\\
    Batch size&32\\
    Learning rate&0.0003\\
    Drop out rate&0.1\\
  \hline
\end{tabular}
\end{table}

To conduct an informative experiment, we rebuild the N-gram model as a single baseline for comparison. Comparison to other approaches~\cite{nguyen2015normalization,nguyen2019ocr,nguyen2019deep} is not conducted due to two main reasons. First, they are proposed for domains different from ours~\cite{nguyen2015normalization,nguyen2019ocr}. Some studies only focus on OCR spelling correction. Apparently, it is not directly comparable to the methods for general solutions like VSEC. Second, some studies primarily focus on a specific sort of Vietnamese spelling errors, such as consonant misspell errors~\cite{nguyen2019deep}. Thus, it is unfair to compare VSEC with methods in these studies. For the above reasons, we evaluate and analyze the current state-of-the-art method, N-gram, to ensure fairness and generality of the experiment.

In addition, the BiLSTM Seq2seq model with attention mechanism and the Transformer models at different token levels are also trained and tested according to the same process. The Vietnamese news corpus is still being used to build these baseline methods.~\subsection{Main Results}

Table~\ref{tb.t4} shows the experimental results of all methods on the test dataset. From the table, we can see that the proposed model substantially outperforms the baseline methods. Particularly, in the detection phase, our proposed method performs much better than the baselines in terms of all metrics. The result for recall of correction task on the test dataset is greater than 76\%, implying that more than 76\% of errors will be fixed.

\setlength{\tabcolsep}{0.5em}
\renewcommand{\arraystretch}{1.3}

\begin{table}[!h]
    \centering{%
    
\caption{Performances of Different Methods on Vietnamese spelling correction}

\label{tb.t4}
    \begin{tabular}{|c|c|c|c|c|c|c|}
    
\hline
 \multirow{2}{*}{\textbf{Method}}& \multicolumn{3}{c|}{\textbf{Detection}} & \multicolumn{3}{c|}{\textbf{Correction}}\\
 
 \cline{2-7}
% \hline
  &\textbf{DP}& \textbf{DR}& \textbf{DF} & \textbf{CP}& \textbf{CR}& \textbf{CF}\\
\hline
N-gram& 0.912 & 0.731& 0.812 & \textbf{0.891} & 0.714 & 0.793 \\
\hline
Seq2seq with attention&0.310 &0.752 & 0.439& 0.222& 0.539& 0.315\\
\hline
\hline
Character-level Transformer& 0.775 & 0.367 & 0.498 & 0.612 & 0.290 & 0.393 \\
\hline
Syllable-level Transformer& 0.719& 0.776 & 0.746 & 0.636 & 0.686 & 0.661 \\
\hline
VSEC & \textbf{0.931} & \textbf{0.813} & \textbf{0.868} & 0.874 & \textbf{0.763} & \textbf{0.815} \\
    
\hline
\end{tabular}}
\end{table}

The N-gram method achieves the highest precision of correction because it can reduce false corrections by using an additional parameter, \textit{error\_threshold}. The use of this parameter is effective with more than 89\% of precision in both evaluation criteria. However, precision and recall are a tradeoff. Increasing the precision of the N-gram method entails lowering the recall. On the other hand, the proposed method shows more balance, when the values of the precision and recall measurements are not significantly different. Specifically, the proposed method reaches 86.8\% with the F1 score measure in the error detection task and 81.5\% in the error correction task, while the N-gram method only reaches 81.2\% and 79.3\%, respectively. 

For the methods of using Seq2seq architecture, the subword-level Transformer model performs better than the other baselines, while the method of the BiLSTM Seq2seq model with attention mechanism performs fairly poorly. This indicates that, despite ignoring traditional recurrent architectures, the Transformer-based models are still able to outperform the LSTM-based models. Furthermore, the subword-level model can beat the models at other token levels. This demonstrates that the subword-level model can handle the out of vocabulary problem better than one at the syllable level and it also performs effectively than the character-level model.

\subsection{Effect of Hyperparameter}

We also investigate the effect of the vocabulary size and the data size.
Table~\ref{tb.t5} shows that the proposed method reaches its best performance with the data size is 5 million. This indicates that the more training data the higher performance can achieve.

% We present the results of the proposed method on the test data to explore the effect of vocabulary size and data size. Table \ref{tb.t5} shows the results of the proposed method learned with different sizes of the training data. One can find that the best result is obtained for the proposed method when the size is 5 million that indicates the more training data the higher performance can achieve.

\setlength{\tabcolsep}{0.5em} % for the horizontal padding
\renewcommand{\arraystretch}{1.3}% for the vertical padding

\begin{table}[!h]
    \centering{%
    
\caption{Impact of Different Sizes of Training Data}

\label{tb.t5}
    \begin{tabular}{|c|c|c|c|c|c|c|}
    
\hline
 \multirow{2}{*}{\textbf{Training Set}}& \multicolumn{3}{c|}{\textbf{Detection}} & \multicolumn{3}{c|}{\textbf{Correction}}\\

\cline{2-7}
   &\textbf{DP}& \textbf{DR}& \textbf{DF} & \textbf{CP}& \textbf{CR}& \textbf{CF}\\
\hline
500K &0.880& 0.679& 0.767& 0.777& 0.599&0.676\\
\hline
1M&  0.896& 0.729 &0.804 & 0.817& 0.665 & 0.733\\
\hline
2M& 0.891 & 0.769 & 0.826 & 0.826 & 0.713 & 0.765\\
\hline
5M & \textbf{0.931} & \textbf{0.813} & \textbf{0.868} & \textbf{0.874} & \textbf{0.763} & \textbf{0.815} \\
    
\hline
\end{tabular}}
\end{table}

A larger vocabulary size means fewer syllables split into two or more tokens. Table~\ref{tb.t6} presents the results of the proposed method in different values of the hyperparameter vocabulary size. The highest F1 score is obtained at the vocabulary size equal to 30,000. That is to say, having a larger vocabulary does not guarantee a higher F1 score.

\setlength{\tabcolsep}{0.5em} % for the horizontal padding
\renewcommand{\arraystretch}{1.3}% for the vertical padding

\begin{table}[!h]
    \centering{%
    
\caption{ Impact of Different Values of Vocabulary size}

\label{tb.t6}
    \begin{tabular}{|c|c|c|c|c|c|c|}
    
\hline
 \multirow{2}{*}{\textbf{Vocabulary size}}& \multicolumn{3}{c|}{\textbf{Detection}} & \multicolumn{3}{c|}{\textbf{Correction}}\\

\cline{2-7}
   &\textbf{DP}& \textbf{DR}& \textbf{DF} & \textbf{CP}& \textbf{CR}& \textbf{CF}\\
\hline
1K& 0.886 & 0.585 & 0.705 & 0.771 & 0.509 & 0.613 \\
\hline
10K& 0.935& 0.793 & 0.858 & \textbf{0.878} &0.745& 0.806 \\
\hline
30K & 0.931 & \textbf{0.813} & \textbf{0.868} & 0.874 & \textbf{0.763} & \textbf{0.815}  \\
\hline
50K & \textbf{0.937} &0.773 & 0.847 & 0.874 & 0.721 &0.790  \\
    
\hline
\end{tabular}}
\end{table}

% \subsection{Decoding Time}
% We used Pytorch in the experiment. This library has a built-in module aimed to conveniently develop a Transformer model. However, because of its accessibility, a major disadvantage of this module is its lack of speed optimization. First, from the formulas (\ref{eqn:eq1}), (\ref{eqn:eq2}), (\ref{eqn:eq3}), it can be seen that the output of the encoder is calculated separately from the decoder. This means that the encoder output results can be calculated once and reused for each step in the decoder.  Second, at the decoding step, a token is generated based only on the tokens of the position in front of it. This implies that instead of computing all token values at each time step, we can cache the previously computed values and only compute the value for the last place token in the time step. Figure \ref{fig:fig5.9} shows the performance of the Transformer model using the caching technique. 

% \begin{figure}[H]
%     \centering
%   \includegraphics[width=350pt]{Figures/cache_machenism.png}
%     \caption{Cache technique performance with 2-core CPU computer (without GPU)}
%     \label{fig:fig5.9}
% \end{figure}

\subsection{Discussion}

  \selectlanguage{vietnamese}
We observed that the proposed method is able to make more effective use of context information than the N-gram method. When there are two or more errors next to each other in a sentence, N-gram usually detects one and leaves the others undetected. The proposed method, on the other hand, can detect both of the errors. For example, there are 2 two errors in the sentence ``Chuẩn bị sãn sang hợp đồng ký kết'' (Prepare the contract to be signed). The syllables ``sãn'' and ``sang'' are incorrect and they should be written as ``sẵn'' ``sàng'', which form a word meaning ``ready''. The N-gram method only corrects one syllable that is ``sãn'' while the proposed method can correct both of them. It is because the N-gram method relies on the context provided by nearby syllables, specifically two syllables before and two syllables after the target.

  \selectlanguage{english}
We also found that the proposed method has three major types of false detections. For statistics of errors, we sampled 100 false detections from the test set. We noticed that 32\% of errors are foreign words and acronym words, 28\% of errors are due to a lack of domain-specific knowledge, and the remaining 40\% of errors have no specific type.

  \selectlanguage{vietnamese}
Foreign words and acronym words are the first type of false detection. These words are sometimes converted to Vietnamese syllables by the model. For example, in the sentence ``TH đã luôn tiếp cận sản xuất theo chuỗi đồng cỏ sạch'' (TH has always approached clean grassland chain production), the acronym word ``TH''  (a Vietnamese company) is converted to ``Thì'' syllable. This indicates that in order to make more reliable detections, the models must have a stronger way to determine what the special syllables are. 

The second type of false detection is due to a lack of domain-specific knowledge. For example, in the sentence ``Đồ thị của hàm số bậc hai'' (Graph of quadratic function). The model turned the word ``Đồ thị'' (Graph) into ``Đô thị'' (City). This happens due to the fact that the test set is inclined to scholarly language while it is not much in the training data created from the news corpus. This problem is still very challenging for the existing model to determine this type of error.

  \selectlanguage{english}

%% file: Sections/conclusions.tex
\section{Conclusions}

\label{conclusions}
In this paper, we propose a neural network approach for Vietnamese spelling correction. Our method is powered by applying a deep learning subword-level model based on Transformer. The technique of subword tokenization is general and potentially useful for dealing with the out-of-vocabulary problem. The Transformer model takes the sequence of subword tokens containing spelling errors as the source and the corrected one as the target. Experimental results on the realistic dataset show that our method outperforms the state-of-the-art model using the N-gram method. For further research, we plan to extend the Error Generator to capture more types of Vietnamese spelling errors and explore pre-trained models such as multilingual BERT~\cite{devlin2018bert} to apply to this task.